\documentclass{sig-alternate-2013}

\newfont{\mycrnotice}{ptmr8t at 7pt}
\newfont{\myconfname}{ptmri8t at 7pt}
\let\crnotice\mycrnotice%
\let\confname\myconfname%

\permission{Permission to make digital or hard copies of all or part of this work for personal or classroom use is granted without fee provided that copies are not made or distributed for profit or commercial advantage and that copies bear this notice and the full citation on the first page. Copyrights for components of this work owned by others than ACM must be honored. Abstracting with credit is permitted. To copy otherwise, or republish, to post on servers or to redistribute to lists, requires prior specific permission and/or a fee. Request permissions from Permissions@acm.org.}
\conferenceinfo{ICMR'15,}{June 23--26, 2015, Shanghai, China.}
\copyrightetc{Copyright \copyright~2015 ACM \the\acmcopyr}
\crdata{978-1-4503-3274-3/15/06\ ...\$15.00.\\
DOI: http://dx.doi.org/10.1145/2671188.2749321}

\usepackage{amsmath}
\usepackage[numbers]{natbib}
\usepackage{array}
\usepackage{algorithmic}
\usepackage{algorithm}

\usepackage{authblk}

\newif\ifeps
\epstrue

\begin{document}


\title{End-to-End Photo-Sketch Generation via Fully Convolutional Representation Learning
}
\renewcommand\Affilfont{\normalfont\normalsize}

\author[1]{\textbf{Liliang Zhang}}
\author[1,*]{\textbf{Liang Lin}}
\author[1]{\textbf{Xian Wu}}
\author[1]{\textbf{Shengyong Ding}}
\author[2]{\textbf{Lei Zhang}}
\affil[1]{Sun Yat-sen University, Guangzhou 510006, China}
\affil[2]{Department of Computing, The Hong Kong
Polytechnic University.}
\affil[ ]{zhangll.level0@gmail.com, linliang@ieee.org}

\renewcommand\Authands{\textbf{,} }


\maketitle
\begin{abstract}
Sketch-based face recognition is an interesting task in vision and multimedia research, yet it is quite challenging due to the great difference between face photos and sketches. In this paper, we propose a novel approach for photo-sketch generation, aiming to automatically transform face photos into detail-preserving personal sketches. Unlike the traditional models synthesizing sketches based on a dictionary of exemplars, we develop a fully convolutional network to learn the end-to-end photo-sketch mapping. Our approach takes whole face photos as inputs and directly generates the corresponding sketch images with efficient inference and learning, in which the architecture is stacked by only convolutional kernels of very small sizes. To well capture the person identity during the photo-sketch transformation, we define our optimization objective in the form of joint generative-discriminative minimization. In particular, a discriminative regularization term is incorporated into the photo-sketch generation, enhancing the discriminability of the generated person sketches against other individuals. Extensive experiments on several standard benchmarks suggest that our approach outperforms other state-of-the-arts in both photo-sketch generation and face sketch verification.
\end{abstract}
\vspace{-2mm}
\vspace{-2mm}


\renewcommand{\thefootnote}{\fnsymbol{footnote}}
\footnotetext[1]{\small Corresponding author is Liang Lin. This work was supported by the Hi-Tech Research and Development Program of China (no.2013AA013801), Guangdong Natural Science Foundation (no.S2013050014548), and the Hong Kong Scholar program.}

\section{INTRODUCTION}
Sketch is an important artistic drawing style and may be the simplest form since it is only composed of lines. An interesting application is searching image databases using free-hand sketch queries \cite{smith1997visualseek, ren2014sketch}.

However, drawing a vivid sketch portrait is time consuming even for a skilled artist. Automatic face sketch generation has been studied for a long time and it has many useful applications for digital entertainment \cite{liu2006mapping}.

Another important application based on face sketch is to assist law enforcement. Assumed that we need to automatically retrieval a photo from the database for a query image, which help the police to narrow down the suspect quickly. Unfortunately, the photo of the suspect maybe unavailable in most cases. To deal with such a problem, the best substitute available is an artist drawing based on the recollection of an eyewitness. 

\begin{figure}[!h]
\centering
\ifeps
\includegraphics[width=0.48\textwidth]{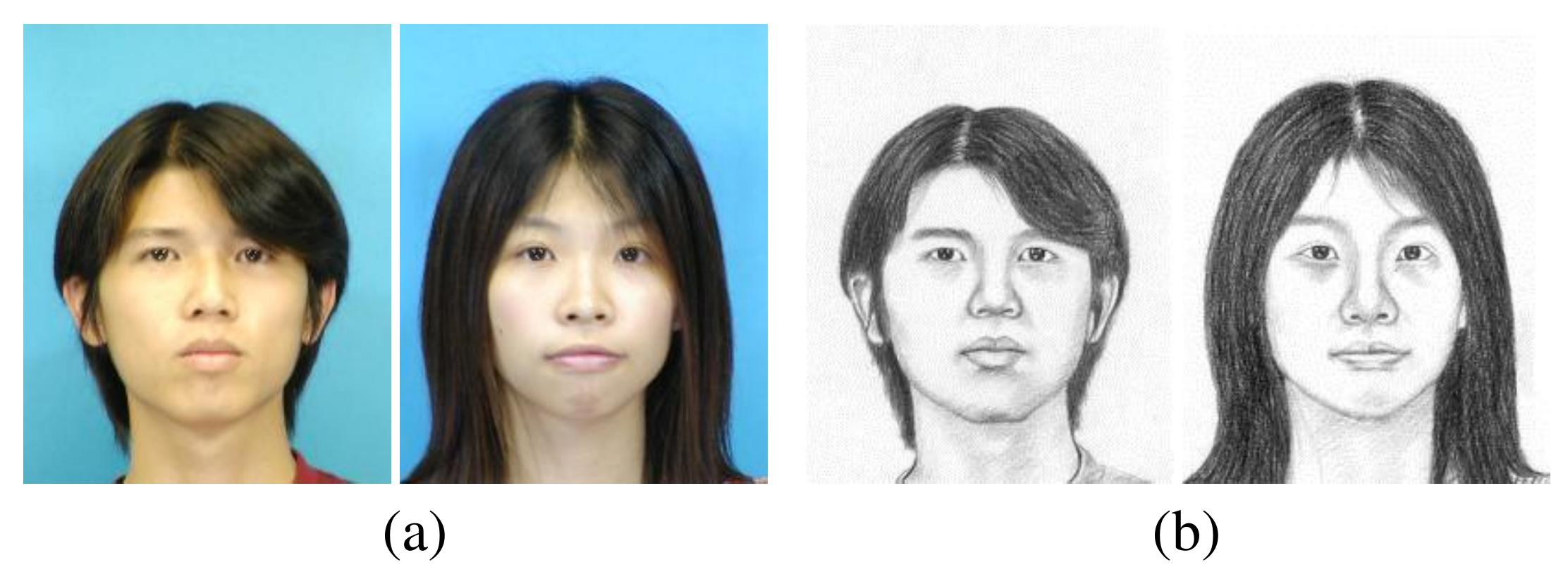}
\else
\includegraphics[width=0.48\textwidth]{figure/00_samples_on_CUFS.jpg}
\fi
\caption{Samples of photo-sketch pairs: (a) photos; (b) sketches drawn by artist}
\label{fig:00_samples_on_CUFS}
\end{figure}

In this situation, we only focus on sketches without exaggeration, so that the sketch can realistically reflect the real person. Figure \ref{fig:00_samples_on_CUFS} shows two samples of photo-sketch pairs in CUHK Face Sketch Database \cite{wang2009face}.

Figure \ref{fig:00_samples_on_CUFS} indicates the great difference between photos and sketches. Thus, photo based face verification methods cannot be directly applied in this problem. The key to sketch based face verification is to reduce the modality difference between photos and sketches.

One intuitive idea is to recover the photo image from a sketch. However, it's an ill-pose problem because a sketch may lose many informations during the drawing procedure. An alternative way is to generate a pseudo-sketch from a photo, which has been discussed in \cite{tang2004face, liu2005nonlinear, wang2009face}. 

The original generation scheme is trying to find a transformation from photos to sketches. Intuitively, it should be a complex nonlinear mapping. Thus, former works like \cite{tang2004face, liu2005nonlinear, wang2009face, xiao2009new, zhang2011svr, wang2012semi, song_eccv14_sketch} simplify the generation problem to \emph{synthesis} form. The underlying
assumption is that, if two photo images (or patches) are similar, their corresponding sketch images (or patches) should also be similar. Thus, if we find out a way to use other photo images (or patches) to synthesize a photo image (or patch), then we can use the corresponding sketch images (or patches) to synthesize its pseudo-sketch.

However, this simplification may has some limitation, especially the scalability. The time cost in synthesis based method grows linearly with the amounts of training data, because it need to use the samples in training set to synthesize a new one.

On the contrary, our method try to \emph{directly} solve the generation problem with an end-to-end model called fully convolutional network (FCN), which can be regarded as a special kind of convolutional neural networks (CNNs). More precisely, FCN is stacked by \emph{only} convolutional layers. It can solve complex nonlinear problem while producing pixel-wise outputs, which is very suitable for the photo-sketch generation problem. Please refer to Figure \ref{fig:06_network_architecture} to get an intuitive idea.

The main contributions of this work are three-folds. First, an end-to-end photo-sketch generation model is studied with a novel architecture of fully convolutional networks, which is original to the best of our knowledge. Second, we present a joint generative-discriminative formulation driving the network optimization, such that the generated face sketches can characterize the person detail as well as the discriminability against other individuals. Third, our system demonstrates superior performances compared with other state-of-the-art approaches on several benchmarks.

Rest of the paper is organized as follows: Section 2 discusses the relative work in photo-sketch synthesis and convolutional neural networks. In Section 3, we will interpret our model and its implementation. In Section 4, extensive experiments suggest that our approach outperforms other state-of-the-art methods on several benchmarks. In Section 5, we draw the conclusion of our work and list some ideas we may adopt in the future.

\section{RELATIVE WORK}

\subsection{Photo-Sketch Synthesis and Recognition}
In recent ten years, there has been several works on the topic of photo-sketch synthesis and recognition. 

\citet{tang2004face} were the first to address the problem with a significant amount of database. They proposed a synthesis method based on eigen transformation (ET). It was under the assumption that the photo-sketch mapping can be approximated as linear, yet this assumption may be too strong especially when the hair region was included. Then, it used reconstruction error based distance for recognition.

\citet{liu2005nonlinear} proposed a nonlinear face sketch synthesis method called locally linear embedding (LLE), which can be considered as an improved version of \cite{tang2004face}. Instead of modelling the whole image, it applied ET on local patches with overlapping. It adopted a kernel-based nonlinear LDA discriminative classifier for sketch recognition.

Another local-based strategy proposed by \citet{xiao2009new} was based on Embedded Hidden Markov model (E-HMM). They transformed the sketches to pseudo-photos and applied eigenface algorithm for recognition.

In \cite{wang2009face}, \citeauthor{wang2009face} proposed a multiscale Markov Random Fields (MRF) for face photo-sketch synthesis, which can be both applied to photo-to-sketch synthesis and sketch-to-photo synthesis. They evaluated a series of classifiers on the pseudo-images. Random Sample LDA (RS-LDA) performed best among them. \citet{zhang2010lighting} then improved the MRF framework by adding shape priors and descriptors robust to lighting variations.

Most of methods above may lose some vital details, which influenced the visual quality and face recognition performance. Thus, \citet{zhang2011svr} added a refinement step on existing approaches. They applied a support vector regression (SVR) based model to synthesize the high-frequency information. Similarly, \citet{gao2012face} proposed a new method called SNS-SRE with two steps, \emph{i.e}.\ sparse neighbor selection (SNS) to get an initial estimation and  sparse-representation-based enhancement (SRE) for further improvement. 

\subsection{Pixel-Wise Predections via CNNs}
Convolutional neural network (CNN) was widely used in computer vision. Its typical structure contains a series of convolutional and pooling layers as feature extractors, and several full connected layers as classifiers to give prediction. It has achieved great success in large scale object classification, localization and detection \cite{krizhevsky2012imagenet, he2014spatial, sermanet-iclr-14, erhan2013scalable, szegedy2013deep, girshick2013rich}. 

One important application of CNN was to produce dense or even pixel-wise predictions. \citet{sermanet-iclr-14} developed a CNN-based framework called OverFeat, which integrated recognition, localization and detection. The key component was a pooling layer with offsets, which can imitate sliding window technique and produce dense outputs in the final layer. A similar idea called spatial pyramid pooling (SPP) was proposed by \citet{he2014spatial}, which was also modified the last pooling layer. \citet{wang2014deep} proposed a joint architecture for generic object extraction, and \citet{luo2013pedestrian} applied a deep decompositional network for pedestrian parsing. Both of them shared an idea, which is adopting a full connected layer on the top of the network to produce a dense prediction. Moreover, patch-by-patch scanning technique on the original image has been studied by \citet{ciresan2012deep} in neuronal membrane segmentation and \citet{farabet2013learning} for scene labeling. 

Recently, \citet{dong2014learning} applied CNN for image super-resolution and it can produce a pixel-wise output. 
They discussed its relationship to sparse-coding-based methods, and concluded that they can use three convolutional layers to simulate the representation-mapping-reconstruction procedure in sparse coding. 

Our work was mostly inspired by \cite{dong2014learning}. We conducted our early experiments via their network and then further improved it. We called it fully convolutional network (FCN, a similar idea was also proposed in \cite{long_shelhamer_fcn}), for it only contains convolutional layers and the corresponding activation function, but without any other layers like pooling, full connected, local response normalization (LRN), \emph{etc}.\ We would further discuss it in Section 3.2.

\section{PHOTO-SKETCH GENERATION}
\subsection{Formulation}
In our model, we use two constraints for sketch generation. The first one is the generated sketches should be as close to the ones drawn by artists as possible. This encourages the deep network to learn the sketching skills from the artists. The second one is the generated sketches should be able to facilitate the law enforcement. That is, given a sketch drawn by the artist, it should be able to identify the subject in the photo database. These two constraints are introduced as generative loss and discriminative regularizer in our objective function, which is similar to \citep{ding2015deep}.

Suppose there are $N$ subjects in the training set with each subject containing one photo $P_i$ and one sketch $S_i$. We use $f(\mathbf W,P_i)$ to denote the sketch generated by a fully convolutional network parameterized by $\mathbf W$. We define our loss function as follows where $L_{gen}$ and $L_{disc rim}$ represent the generative loss and the discriminative regularizer respectively. Here we use $\alpha$ to control the weight of the discriminative regularizer to the overall objective. 
\begin{equation}
L(P,S,\mathbf W)=L_{gen}(P,S,\mathbf W)+\alpha L_{discrim}(P,S,\mathbf W)
\end{equation}

For the generative loss, we use a straight function which is defined as the pixel-wise difference between the ground-truth sketch and the generated one. 
\begin{equation}
L_{gen}\left(P, S, \mathbf W \right) = \frac{1}{N}\sum_{i=1}^{N}\left ( S_{i} - f(\mathbf W,P_{i}) \right ) ^ 2 
\label{equation:generative-loss}
\end{equation}

For the discriminative regularizer, we encourage the drawn sketch of one particular person should be different from the generated sketch of another person as defined by the following function. 
\begin{equation}
\begin{split}
&L_{discrim}\left(P, S, \mathbf W \right) = \\
&\frac{1}{N\left(N-1 \right)}\sum_{i = 1}^{N}\sum_{j=1,j\not= i}^{N}\log\left(1 + \mathrm{e} ^{ - \frac{\left(S_{i} -  f(W,P_{j})\right)^2}{	\lambda}} \right)
\end{split}
\label{equation:discriminative_loss}
\end{equation}

$\lambda$ is a parameter used to avoid numeric overflow.

\textbf{Parameter Optimization}
Using this model, the learning process is to minimize the loss function $L(P,S,\mathbf W)$. This is achieved by the standard network propagation algorithm in the batch training manner. More precisely, in each iteration, we randomly select a set of photo-sketch pairs and construct the generative loss items and discriminative regularizer items respectively. In order to derive the gradient with respect to the network parameter $\mathbf W$,  the key step is to calculate the partial derivative of the loss function with respect to the output (generated sketch) of each photo. As one photo may be involved into several cost items, we need to go through each items to accumulate the derivative with respect to the output. Algorithm $\ref{alg:batchTraining}$ gives the details of the learning process. 
\begin{small}
\begin{algorithm}[htb]
\caption{Parameter Optimization}
\label{alg:batchTraining}
\begin{algorithmic}[1]
\REQUIRE ~~\\
   Training photos $\{P_i\}$ and sketches $\{S_i\}$;
\ENSURE ~~\\
	Network Parameters $\mathbf{W}$
\WHILE    {$t<T$}
\STATE $t\leftarrow t+1$;
\STATE Randomly select a subset of photos and sketches $\{P'_i\},\{S'_j\}$ from the training set;
\FORALL {$P'_i$}
\STATE Do forward propagation to get $f(\mathbf W, P'_i)$
\ENDFOR
\STATE $\Delta \mathbf W=0$
\FORALL {$P'_i$}
\STATE Calculate the partial derivative with respect to the output: $\frac {\partial L}{\partial f(\mathbf W, P'_i)}$ 
\STATE Run backward propagation to obtain the gradient with respect to the network parameter: $\Delta \mathbf W_i$
\STATE Accumulate the gradient: $\mathbf W+=\Delta \mathbf W_i$
\ENDFOR
\STATE $\mathbf{W}^t=\mathbf{W}^{t-1}-\lambda_t \Delta \mathbf{W}$
\ENDWHILE

\end{algorithmic}
\end{algorithm}
\end{small}

\begin{figure*}[!t]
\centering
\ifeps
\includegraphics[width=1.0\textwidth]{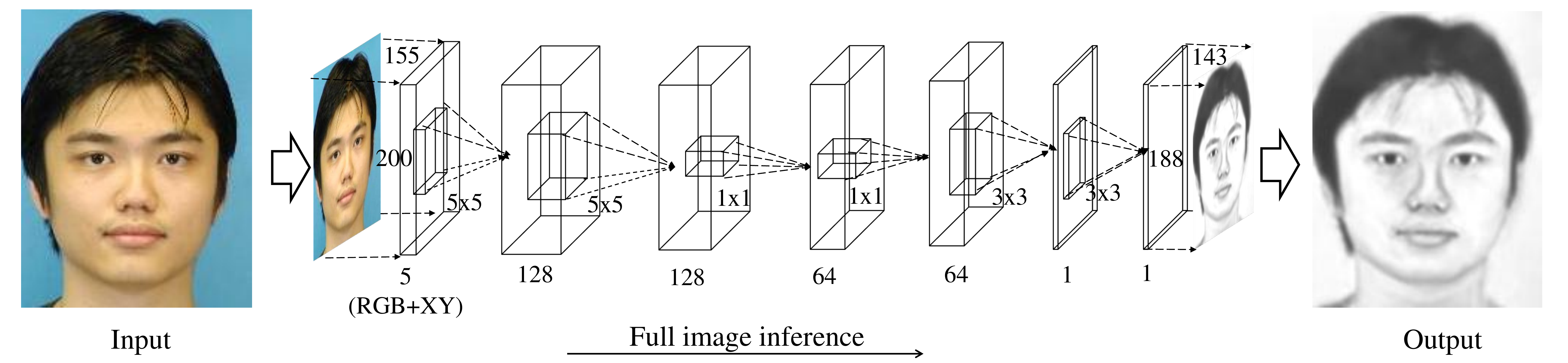}
\else
\includegraphics[width=1.0\textwidth]{figure/06_network_architecture.jpg}
\fi
\caption{The overview of our model. It takes a full-size photo image as input and directly generates a full-size pseudo-sketch as output. The middle part is the architecture of our fully convolutional network. It contains six convolutional layers, with rectified linear units as activation functions (omitted in the figure).}
\label{fig:06_network_architecture}
\end{figure*}

\subsection{Fully Convolutional Representation}
In this subsection, we will discuss the property of convolutional layers and how they can be cascaded as a pixel-wise fully convolutional representation of an input image.

\textbf{Composition of Convolutions}
A typical convolutional layer has $K$ kernels with activation function $f$ can be formulated as 
\begin{equation}
y^{k}_{ij} = f((\mathbf{W}_{k} * x)_{ij}+b_{k})
\label{equation:convolutional layer}
\end{equation}

$x$ denotes the input feature maps. $k \in \{1,2,...,K\}$ denotes the index of a filter, and $\mathbf{W}_{k}$ and $b_{k}$ are the $k$-th filter weights and bias. $y^{k}_{ij}$ denotes the element at the coordinate $(i,j)$ on the $k$-th output feature map.

Equation (\ref{equation:convolutional layer}) indicates that the convolutional operation preserves the spatial relationship. What's more, composition of convolutions will not change this property, \emph{i.e}.\ we can use a stack of convolutional layers to represent a complex nonliear mapping, which can be adopted on an input of arbitrary size, and then produce a corresponding spatial output.

\textbf{Relationship to the Patch-Wise Representation}
In Equation (\ref{equation:convolutional layer}), a pixel on the output feature map only will be influenced by a \emph{patch} on the input feature map, \emph{i.e}.\ its receptive field. This property is also maintained under composition of convolutions. Thus, fully convolutional representation on the whole image can be regarded as a \emph{batch} of patch-wise representation on patches of the image. Nevertheless, fully convolutional representation is much more efficient, for the patches overlap significantly (we set stride to 1 in all layers).

\textbf{Border Effect Trade-Off}
Convolutional operations will shrink the size of the feature map. For example, if we adopt a $(3 \times 3)$ convolution operator on a $(3 \times 3)$ input, the output size will be shrank to $(1 \times 1)$.

As more convolutional layers stacked up, more shrinks will be accumulated in the outputs. A possible solution is to add proper padding before the convolution operation. But it will bring on \emph{border effects}, which has been claimed in \cite{dong2014learning}. 

Thus, in the trade-off, we don't use padding in the convolutional layers during both the training and testing time. Each $(155\times 200)$ photo image will be shrank to $(143 \times 188)$ in the representation (See Figure \ref{fig:06_network_architecture}). 

\subsection{Implementation}

\textbf{Pre-processing}
As it is described in \cite{wang2009face}, all the photos and sketches are translated, rotated, and scaled such that the two eye centers of all the face images are at fixed position in the pre-processing step. This simple geometric normalization step makes  the same face components in different images in roughly alignment.

Another pre-processing step is inspired by the results\footnote{http://www.ee.cuhk.edu.hk/\~{}xgwang/sketch\_multiscale.html} of \cite{wang2009face}. The transformation mentioned above produces a $(200\times250)$ image, but it may have some black regions on the border areas. We crop the $(155\times200)$ center part of the image in order to exclude this negative influence. 

Since we choose to avoid border effects, the sketch images need to be cropped to $(143 \times 188)$ to fit in the network output dimension. 

\textbf{Spatial Patch-wise Learning with Overlapping}
From the previous works on photo-sketch synthesis\cite{liu2005nonlinear, wang2009face}, patch-wise learning with overlapping is very important to handle the non-linearity between photos and sketches. Intuitively, patches in different positions are diverse from others (\emph{e.g}.\ eye patches, nose patches and mouth patches). Therefore, learning different patch representations in different spatial positions respectively is a very straightforward idea. 

We handle it by adding additional XY channels in the input, \emph{i.e}.\ the input image data contain five channels, three RGB channels of the photo image, and two channels of the corresponding coordinate $(i, j)$. This tiny modification significantly improves the result, which will be discussed in Section 4.3.

\textbf{Network Architecture}
We apply the network proposed in \cite{dong2014learning} in our early experiments, then we modify its architecture for further enhancement. We borrow the idea in \cite{simonyan2014very}, which has a great success in ILSVRC-2014. Their main contribution is  to deeper the CNN under computational constraints via very
small $(3\times3)$ convolutional filters in all layers.

Similarly, we modify the 3 layers network in \cite{dong2014learning} into a 6 layers network, which is shown in Figure \ref{fig:06_network_architecture}. For the $(9\times9)$, $(1\times1)$, $(5\times5)$ convolutional layer in \cite{dong2014learning}, we replace them with two $(5\times5)$, $(1\times1)$ and $(3\times3)$ convolutional layers respectively. Moveover, we double the filter amounts in every layer to improve the network's capacity.

We call it \emph{medium network} due to its size. We have two similar architectures called small network and large network, which will be further discussed in Section 4.3.

\textbf{Training Details}
Our results are based on our medium network (See Figure \ref{fig:06_network_architecture}). 
As mentioned above, we first pre-process the photo images into $(155\times200)$ and the sketch images into $(143\times188)$. The optimization objective has been discussed in Section 3.1. 
We set $\lambda$ as $10^9$ and set $\alpha$ as $10^4$.

We use Caffe \cite{jia2014caffe} for implementation.
The filter weights are initialized by drawing randomly from a Gaussian distribution with zero mean and standard deviation 0.01, and the bias are initialized by zero. We set the learning rate as $10^{-11}$, then it takes several hours to converge on a NVIDIA Tesla K40 GPU. 

\begin{figure*}[!t]
\centering
\ifeps
\includegraphics[width=1.0\textwidth]{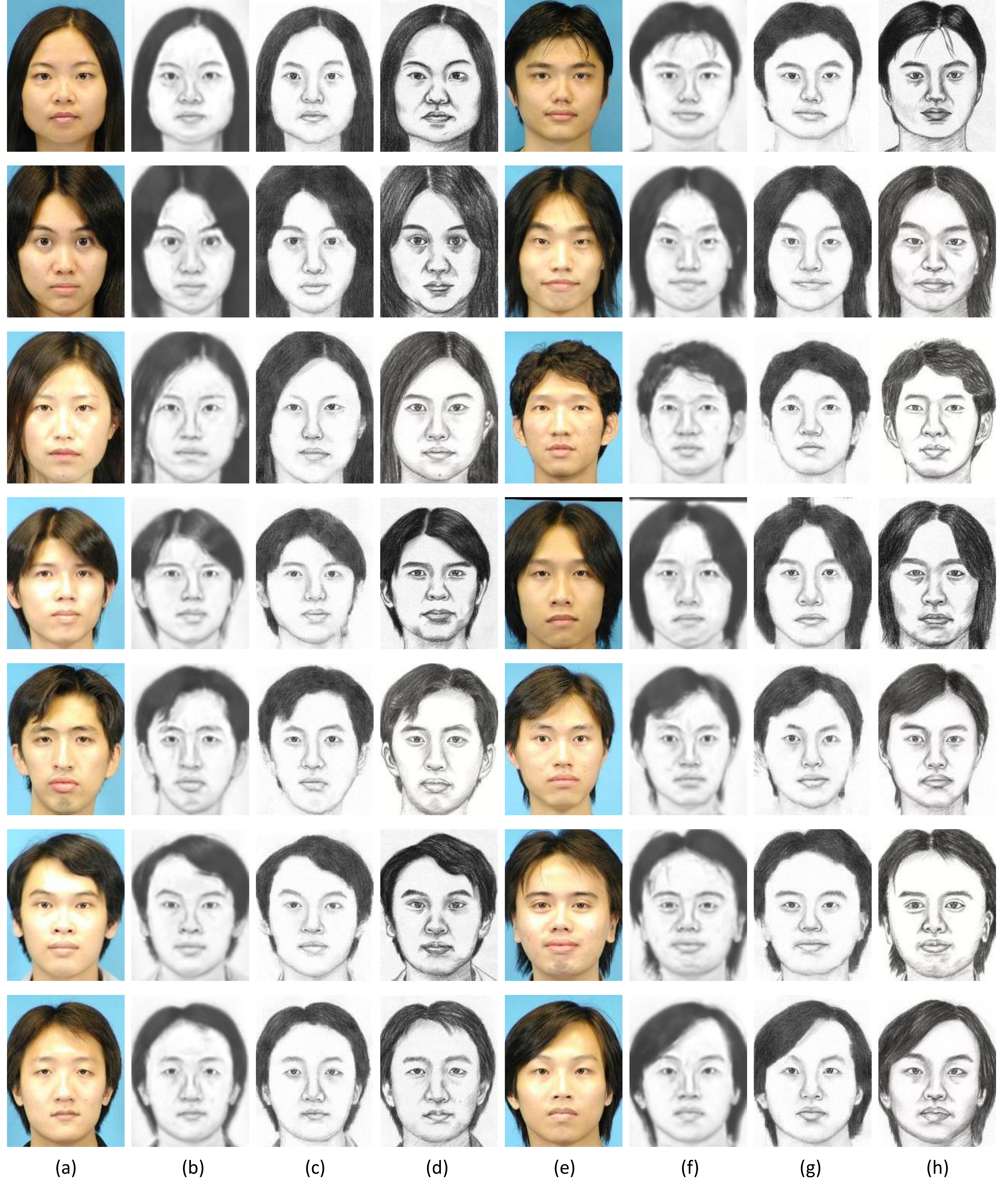}
\else
\includegraphics[width=1.0\textwidth]{figure/01_photo_to_sketch_fig.jpg}
\fi
\caption{Compared to synthesis based method \cite{wang2009face}, our generative approach could retain more details in the original photos, \emph{e.g}.\ the two locks of hair in forehead of the man in column (e) of row 1; the two big eyes of the woman in column (a) of row 2. (a)\&(e) photos; (b)\&(f) our generative pseudo-sketches; (c)\&(g) synthesized pseudo-sketches in \cite{wang2009face}; (d)\&(h) sketches drawn by the artist.  }
\label{fig:01_photo_to_sketch_fig}
\end{figure*}

\section{EXPERIMENTS}
We evaluate our model on CHUK student dataset \cite{wang2009face}. It includes 188 faces\if false and can be downloaded from this website (http:/mmlab.ie.cuhk.edu.hk/facesketch.html)\fi. 88 faces are selected for training and remaining 100 faces are selected for testing. For each face, there is a sketch drawn by the artist and a photo taken in a frontal pose, under normal lighting condition, and with a neural expression. 

We mainly do three aspects of experiments, which will be shown in the next three subsections. Section 4.1 will discuss our photo-sketch generation results and comparison with the synthesis based method in \cite{wang2009face}. Section 4.2 will show that our generated pseudo-sketches significantly reduce the modality between photos and sketches. Thus they can be adopted to a sketch-based face verification system. Section 4.3 will involve empirical study about our model. Several factors such as network depth and filter numbers will be discussed on the evaluation protocol both qualitatively and quantitatively.

\subsection{Face-Sketch Generation}
In Figure \ref{fig:01_photo_to_sketch_fig}, we show some examples of our generated results and compare them to \cite{wang2009face}. For fair comparison, we list all 14 pseudo-sketches which can be found on their website.

Figure \ref{fig:01_photo_to_sketch_fig} indicates that our results contain more vital details.
For example, the man in column (e) of the first row has two little locks of hair on his forehead. But this detail is missing in the pseudo-sketch in \cite{wang2009face} (in column (g)). On the contrary, our generated result (in column(f)) retains this detail information which may be very important to distinguish this man from others. 
Another example is the woman in column (a) of the second row. She has a distinctive feature comparing to other persons in the database, for she has two big eyes. Our method can also capture this detail in the pseudo-sketch (in column(b)).

It suggests that synthesis based methods may fail in some cases. The main reason is that they are under an assumption that the synthesized sketch image (or patch) can be reconstructed by the sketch images (or patches) in the training set. However, this assumption may be too strong, for some persons have their facial distinctions in fact.

Our generative method overcomes this difficulty.
We try to \emph{directly} learn the transformation from photo space to sketch space. Even though it's a complex nonlinear mapping, FCN has the capacity to handle this challenge.

Compared to traditional synthesized method, our novel generative approach has two folds of advantages. Firstly, it could retain more detail information from the photos. Secondly, our inference time is independent on the amount of training data. The time cost in synthesis based method grows linearly with the data amounts, while the runtime of our approach is only affected by the size of the input photo.  

  
\subsection{Sketch-Based Face Verification}
In this subsection, we will show that our generated pseudo-sketch significantly reduce the modality between photos and sketches. We follow the same testing procedure in \cite{tang2004face, liu2005nonlinear, wang2009face}, which can be concluded in two steps: (a) convert the photos in testing set into corresponding pseudo-sketches; (b) define a feature or transformation to measure the distance between the query sketch and the pseudo-sketches.  

In our implementation, we use our model to generate the pseudo-sketches in procedure (a), and use our generative loss (see Equation (\ref{equation:generative-loss})) as the distance measurement. Since the sizes of our pseudo-sketches are $(188\times143)$, we also crop the query sketch into $(188\times143)$. 

Following the same protocol described in \cite{tang2004face}, we compare our approach with previous methods on cumulative match score ($CMS$) in Table \ref{table:CMS-comparison-with-others}. $CMS$ measures the percentage of ``the correct answer is in the top $n$ matches'', where $n$ is called the rank.


As shown in Table \ref{table:CMS-comparison-with-others}, we form a baseline experiment, which is to convert the photo into gray scale as somehow ``pseudo-sketch'' to clarify the modality difference between photo space and sketch space. In this baseline, the accuracy for the first rank only equals to 41\%, which is far away from satisfying. In early method like ET described in \cite{tang2004face}, has got 71\% for the top one match. Latter trials in \cite{wang2009face}, \cite{zhang2010lighting}, \cite{zhang2011svr} have greatly improved the verification accuracy on top one candidate to 96\%, 99\% and even 100\%.

From the last row in Table \ref{table:CMS-comparison-with-others}, our approach also gets 100\% correct answer on its first guess. It's worth to notice that our approach is quite different from the former methods, for it applies generation instead of synthesis. 

\begin{table}[!h] 
\center
\begin{tabular}{|>{\hfil}p{50pt}<{\hfil}|>{\hfil}p{32pt}<{\hfil}|>{\hfil}p{32pt}<{\hfil}|>{\hfil}p{32pt}<{\hfil}|>{\hfil}p{35pt}<{\hfil}|}
\hline
       & Rank 1 & Rank 3 & Rank 5 & Rank 10\\
\hline
    Baseline & 41 & 56  & 59  & 70\\
\hline
    ET \cite{tang2004face} & 71 & 81 & 88 & 96 \\ 
\hline
    MRF \cite{wang2009face} & 96 &  - & - & 100 \\ 
\hline
    MRF+ \cite{zhang2010lighting} & 99 & - & - & 100 \\ 
\hline
    SVR \cite{zhang2011svr} & 100 & - & - & - \\ 
\hline
    Ours & \textbf{100} & \textbf{100} & \textbf{100} & \textbf{100} \\
\hline
\end{tabular}
\caption{Cumulative Match Scores ($CMS$) comparison on full training set (88 photo-sketch pairs). Our method achieves 100\% accuracy on the first guess.}
\label{table:CMS-comparison-with-others}
\end{table}

\begin{figure}[!h]
\ifeps
\includegraphics[width=0.48\textwidth]{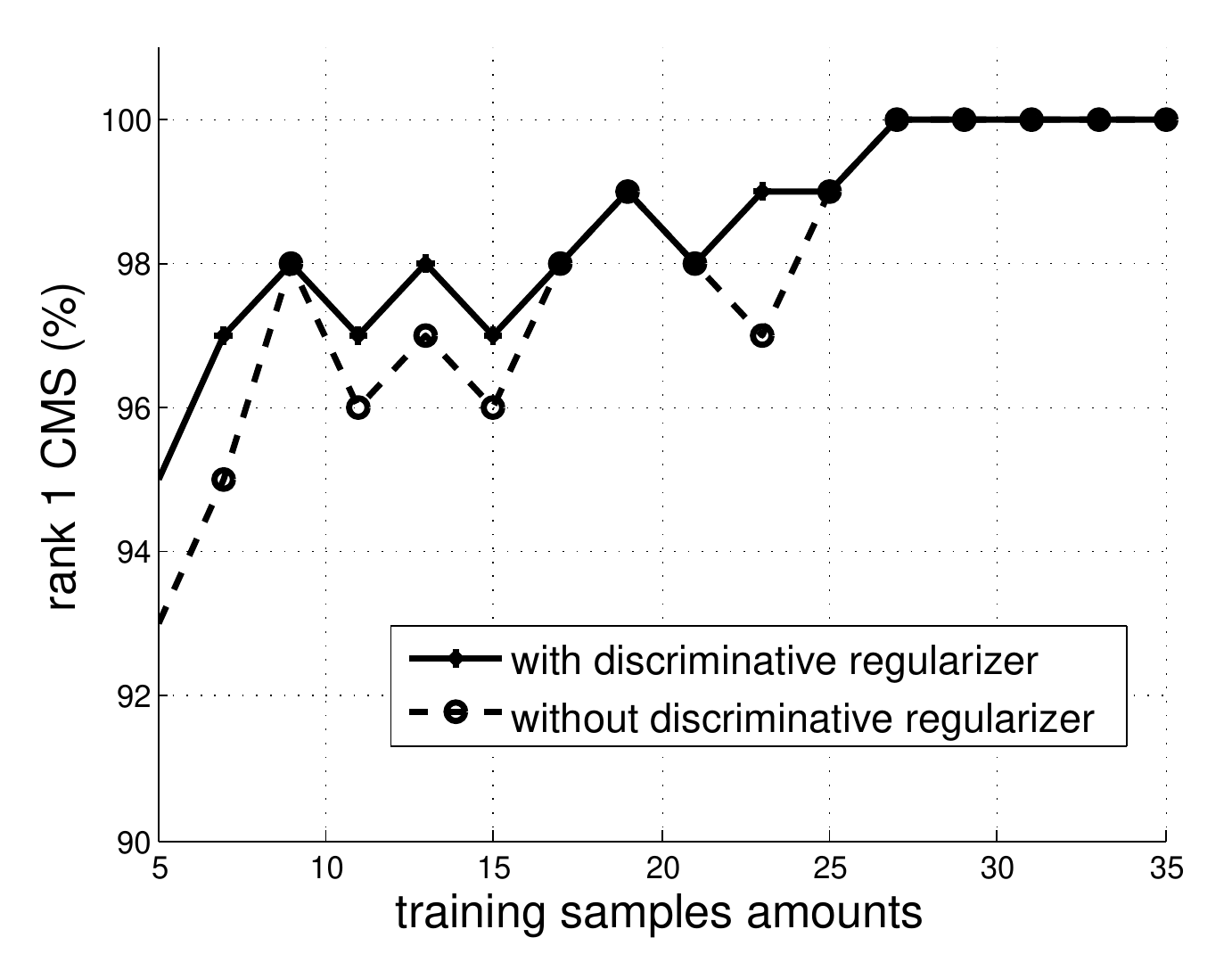}
\else
\includegraphics[width=0.48\textwidth]{figure/07_icmr_discriminative_loss.jpg}
\fi
\caption{The model trained with discriminative regularizer consistently outperforms the model trained without discriminative regularizer. Both model gets 100\% accuracy on rank 1 candidate while involves less than half training samples. \emph{i.e}.\ 27 samples in 88 photo-sketch pairs.}
\label{fig:07_icmr_discriminative_loss}
\end{figure}

Moreover, our generative method is very robust and has excellent generalization ability. In Figure \ref{fig:07_icmr_discriminative_loss}, it suggests that our method can using a portion of training data to get a 100\% accuracy for the first rank. For example, we only need 5 training samples to get to 95\% accuracy, and 27 training samples is enough to get a 100\% score for top one candidate.

We also design a verification on our optimization objective on Figure \ref{fig:07_icmr_discriminative_loss}. It suggests that the model trained with the discriminative regularizer consistently outperforms the model trained without discriminative regularizer.

\subsection{Empirical Study}
In this subsection, further analyses will be conducted on the factors which affect our photo-sketch generation performance.

\textbf{Different Network Architectures} 
Our first concern is about the depth of the network and the numbers of filters. We use the network proposed in \cite{dong2014learning} (we call it SR network) in our early experiments. However, the result is not quite satisfying, for this network seems too small to learn the complex nonlinear mapping from photos to sketches. Thus, we improve the network by: (a) adding more layers to the network; and (b) adding more numbers of filters to each layer.

We conduct our experiments on three different network architectures. Due to their scales, we call them small, medium and large network respectively.

The medium network architecture can be found in Figure \ref{fig:06_network_architecture}. The difference between it and SR network please refer to Section 3.3. Moreover, the only difference between our small, medium and large network is their filter numbers. The medium network has two times of filters compared to the small one. And the large network doubles the filter numbers of the medium network. 

%

For comparison, we define a measurement call $MPRL$, which is short for \emph{multiscale pixel-wise reconstruction loss}. It evaluates the pixel-wise accuracy on different scales. To explain $MPRL$, we firstly introduce \emph{pixel-wise reconstruction loss} ($PRL$) on one photo-sketch pair.

For a photo-sketch pair, we denote the sketch as the ground truth ($GT$) and generated pseudo-sketch as the prediction ($P$). Both of their sizes are $(W \times H)$. Thus, $PRL$ can be formulated as
\begin{equation}
PRL = \frac{1}{{W \times H}} \sqrt{ \sum\limits_{x = 1}^W {\sum\limits_{y = 1}^H { \left(GT(x,y) - P(x,y) \right)^2 } } }
\end{equation}

$MPRL$ is the multiscale version of $PRL$. In practice, for each photo-sketch pair, we rescale both the $GT$ and the $P$ to three scales \{0.5, 1, 2\} and evaluate their $PRL$ to form the $MPRL$. Then, for the whole test set, we evaluate all the pairs and average their $MPRL$ as the final measurement.

\begin{figure}[!t]
\centering
\ifeps
\includegraphics[width=0.49\textwidth]{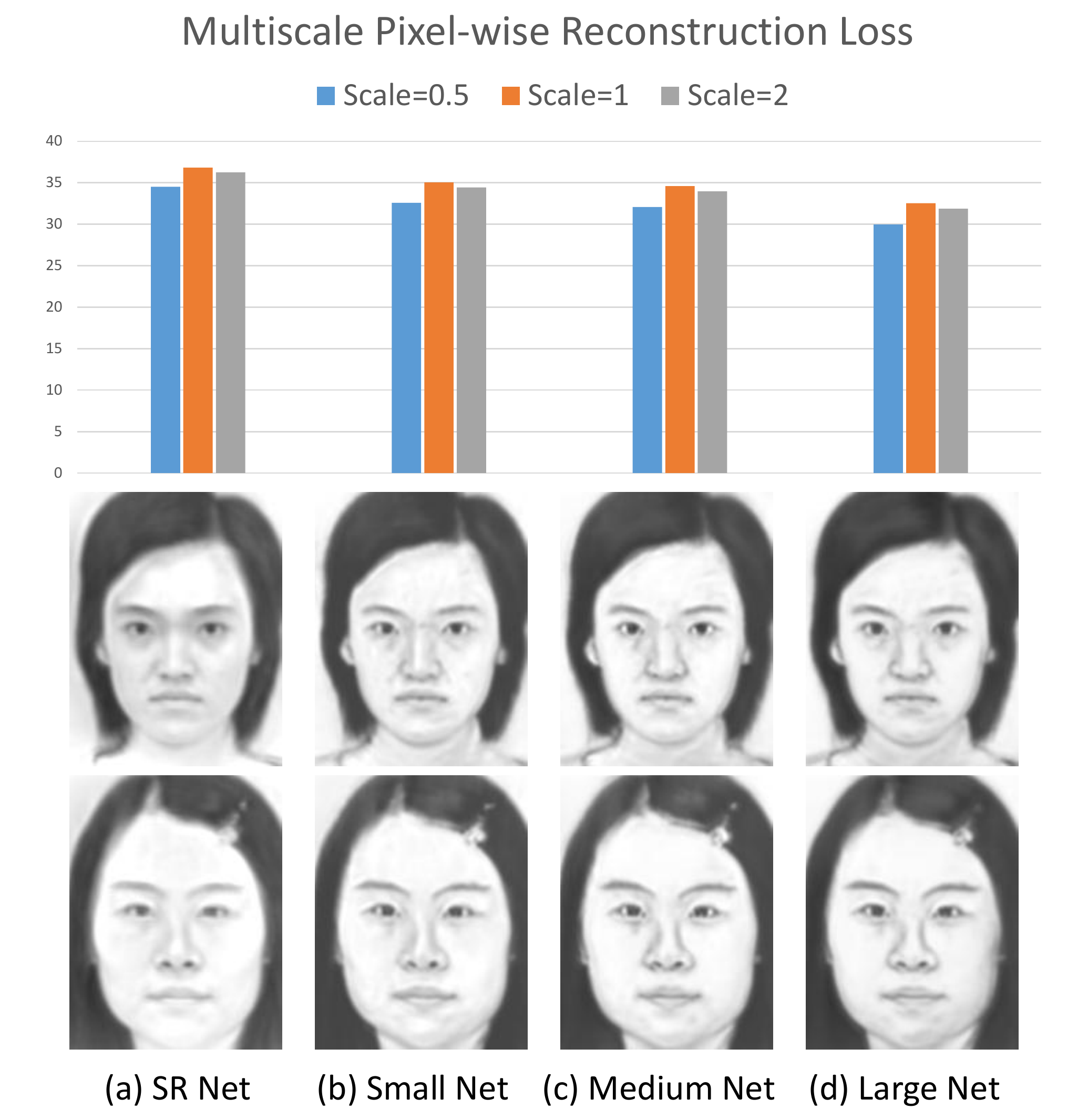}
\else
\includegraphics[width=0.49\textwidth]{figure/02_small_medium_big.jpg}
\fi
\caption{Comparison on the $MPRL$ measurement and generative pseudo-sketches via different network architectures. The model achieves better performance while to deeper the network or to add more filters. (a-d) pseudo-sketches generated via network in \cite{dong2014learning}, our small network, our medium network and our large network.}
\label{fig:02_small_medium_big}
\end{figure}

Figure \ref{fig:02_small_medium_big} summarizes the results both in qualitative and quantitative. The $MPRL$ of SR network is \{34.5, 36.8, 36.2\}. Our small network reduces it to \{32.6, 35.0, 34.4\}. Then, our medium and large network get better performance as \{32.1, 34.6, 34.0\} and \{30.0, 32.5, 31.9\}. 

The pseudo-sketch examples generated by each network are on the bottom part of Figure \ref{fig:02_small_medium_big}. It suggests that the results of SR network can capture the overall structures, but they are far away from satisfying in details. Then, Our small network's generation is quite better and has clearer contours. Moreover, the pseudo-sketches generated by medium network are more vivid than the small ones. 
The results of large network are as good as medium ones, but the large network is much more time consuming.

Table \ref{table:run-time-on-different-networks} shows that our small network is almost as fast as the SR network. The medium network's cost is about 2 times of the small one, and the large network's runtime is about three times of the medium one.

To balance the effectiveness and efficiency, we prefer our medium network as the final choice.

\begin{table}[!h] \normalsize
\center
\begin{tabular}{|>{\hfil}p{22pt}<{\hfil}|>{\hfil}p{30pt}<{\hfil}|>{\hfil}p{40pt}<{\hfil}|>{\hfil}p{52pt}<{\hfil}|>{\hfil}p{41pt}<{\hfil}|}
\hline
      & SR Net & Small Net & Medium Net &  Large Net\\
\hline
   Time & 8.5ms & 8.6ms & 17.1ms & 50.8ms \\
\hline
\end{tabular}
\caption{Runtime of single $(155\times 200)$ image on NVIDIA Tesla K40 GPU}
\label{table:run-time-on-different-networks}
\end{table}

\textbf{Difference between trained with and without XY channels}  
Now we consider another factor that influences the model's performance. As mentioned above, we add XY channels to the RGB color channels. We assume these additional spatial messages could help the network to distinguish different spatial patches and learning different representation on them.

\begin{figure}[!hb]
\centering
\ifeps
\includegraphics[width=0.49\textwidth]{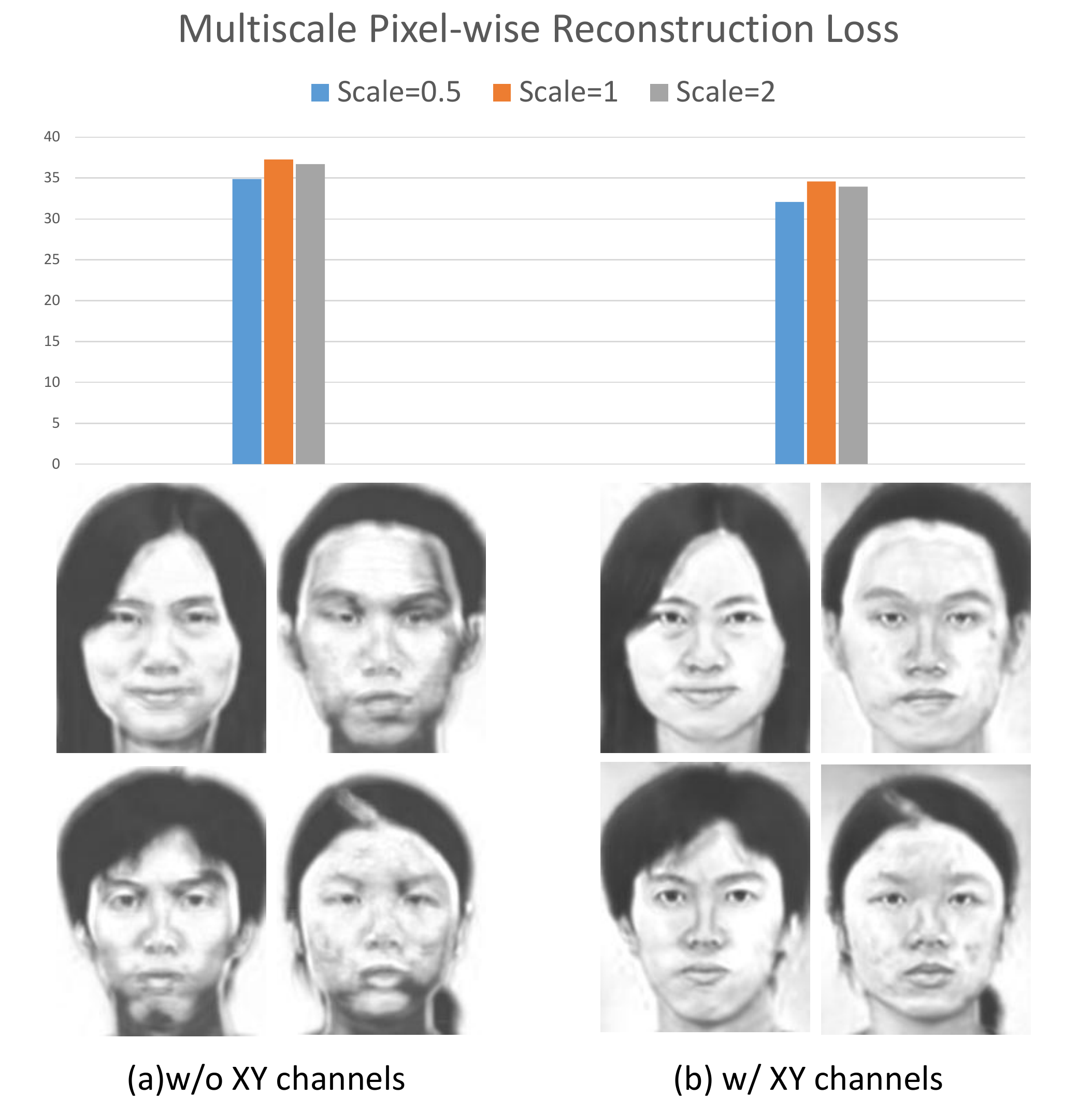}
\else
\includegraphics[width=0.49\textwidth]{figure/03_xychannel.jpg}
\fi
\caption{Comparison on the $MPRL$ measurement and generative pseudo-sketches via network trained without and with XY channels. Adding XY channels can make the model more robust. (a) pseudo-sketches generated via network trained without XY channels; (b) pseudo-sketches generated via network trained with XY channels.}
\label{fig:03_xychannel}
\end{figure}

Similarly, we use $MPRL$ for quantitative measurement and use pseudo-sketch comparison for qualitative analysis. In Figure \ref{fig:03_xychannel}, the left part is for results generated by our medium network but trained without XY channels, and the right part is for results generated by our medium network trained normally. Figure \ref{fig:03_xychannel} suggests that latter one outperforms former one both in numbers and perception. 

\section{CONCLUSION AND FUTURE WORK}
In this paper, we propose an end-to-end fully convolutional network in order to directly model the complex nonlinear mapping between face photos and sketches. From the experiments, we find out that fully convolutional network is a powerful tool which can handle this difficult problem while providing a pixel-wise prediction both effectively and efficiently.

However, this solution still has some limitations. Firstly, the synthesized pseudo-sketches in \cite{wang2009face} have shaper edges and clearer contours than our generated ones. Secondly, the database involved in our experiments only contains Asiatic face images, which may limit the generalization ability of our model to other racial groups. 

In future work, we will further improve our loss function and try various databases in our experiments, and we may explore about the relation between our work and those involved with non-photorealistic rendering \cite{lin2013video}.

\textbf{Acknowledgements}
We gratefully acknowledge NVIDIA for GPU donations.


\setlength{\bibsep}{0.5ex}
\bibliographystyle{abbrvnat}

\bibliography{sketch}


\end{document}